\documentclass[letterpaper]{article} 
\usepackage{aaai25}  
\usepackage{times}  
\usepackage{helvet}  
\usepackage{courier}  
\usepackage[hyphens]{url}  
\usepackage{graphicx} 
\usepackage{natbib}  
\usepackage{caption} 
\usepackage{algorithm}
\usepackage{algorithmic}

\usepackage{algorithm}
\usepackage{algorithmic}
\usepackage{multirow,booktabs}
\usepackage{pifont}
\usepackage{makecell}
\usepackage{dingbat}
\usepackage{amssymb}
\usepackage{amsmath}

\usepackage{newfloat}
\usepackage{listings}
\DeclareCaptionStyle{ruled}{labelfont=normalfont,labelsep=colon,strut=off} 
\floatstyle{ruled}
\newfloat{listing}{tb}{lst}{}
\floatname{listing}{Listing}
\title{Bootstraping Clustering of Gaussians for View-consistent 3D Scene Understanding}
\author {
    Wenbo Zhang\textsuperscript{\rm 1},
    Lu Zhang\textsuperscript{\rm 1}\thanks{Corresponding author},
    Ping Hu\textsuperscript{\rm 2},
    Liqian Ma\textsuperscript{\rm 3},
    Yunzhi Zhuge\textsuperscript{\rm 1},
    Huchuan Lu\textsuperscript{\rm 1}
}
\affiliations {
    \textsuperscript{\rm 1}Dalian University of Technology\\ 
    \textsuperscript{\rm 2}University of Electronic Science and Technology of China\\
    \textsuperscript{\rm 3}ZMO AI\\
    zwbo@mail.dlut.edu.cn, zhangluu@dlut.edu.cn
}

\begin{document}

\maketitle

\begin{abstract}
Injecting semantics into 3D Gaussian Splatting (3DGS) has recently garnered significant attention. While current approaches typically distill 3D semantic features from 2D foundational models (e.g., CLIP and SAM) to facilitate novel view segmentation and semantic understanding, their heavy reliance on 2D supervision can undermine cross-view semantic consistency and necessitate complex data preparation processes, therefore hindering view-consistent scene understanding.
In this work, we present \textbf{\textit{FreeGS}}, an unsupervised semantic-embedded 3DGS framework that achieves view-consistent 3D scene understanding without the need for 2D labels. Instead of directly learning semantic features, we introduce the IDentity-coupled Semantic Field (IDSF) into 3DGS, which captures both semantic representations and view-consistent instance indices for each Gaussian. We optimize IDSF with a two-step alternating strategy: semantics help to extract coherent instances in 3D space, while the resulting instances regularize the injection of stable semantics from 2D space. Additionally, we adopt a 2D-3D joint contrastive loss to enhance the complementarity between view-consistent 3D geometry and rich semantics during the bootstrapping process, enabling \textbf{\textit{FreeGS}} to uniformly perform tasks such as novel-view semantic segmentation, object selection, and 3D object detection.
Extensive experiments on LERF-Mask, 3D-OVS, and ScanNet datasets demonstrate that \textbf{\textit{FreeGS}} performs comparably to state-of-the-art methods while avoiding the complex data preprocessing workload. 
\end{abstract}

%
\begin{links}
    \link{Code}{https://github.com/wb014/FreeGS}
\end{links}

\section{Introduction}
Recently, there has been a surge in the research of 3D scene representation techniques like Neural Radiance Fields (NeRF) \cite{tschernezki2022neural} and 3D Gaussian Splatting (3DGS) \cite{kerbl20233DGS}, owing to their remarkable ability to represent 3D scenes from multi-view images. This trend has spurred the development of approaches that extend the capabilities of radiance fields beyond mere scene reconstruction, enabling semantic-aware tasks such as object segmentation and editing. Pioneering methods \cite{kobayashi2022decomposing,kerr2023lerf,liu2023weakly} have integrated learnable semantic fields into NeRF, performing 3D feature distillation from 2D foundational models \cite{CLIP,li2022language}. However, the introduction of these high-dimensional semantics increases the computational complexity of the original rendering process, adding further burdens to both learning and inference with NeRF's implicit radiance field.

\begin{figure}[t]
  \centering
   \includegraphics[width=1.0\linewidth]{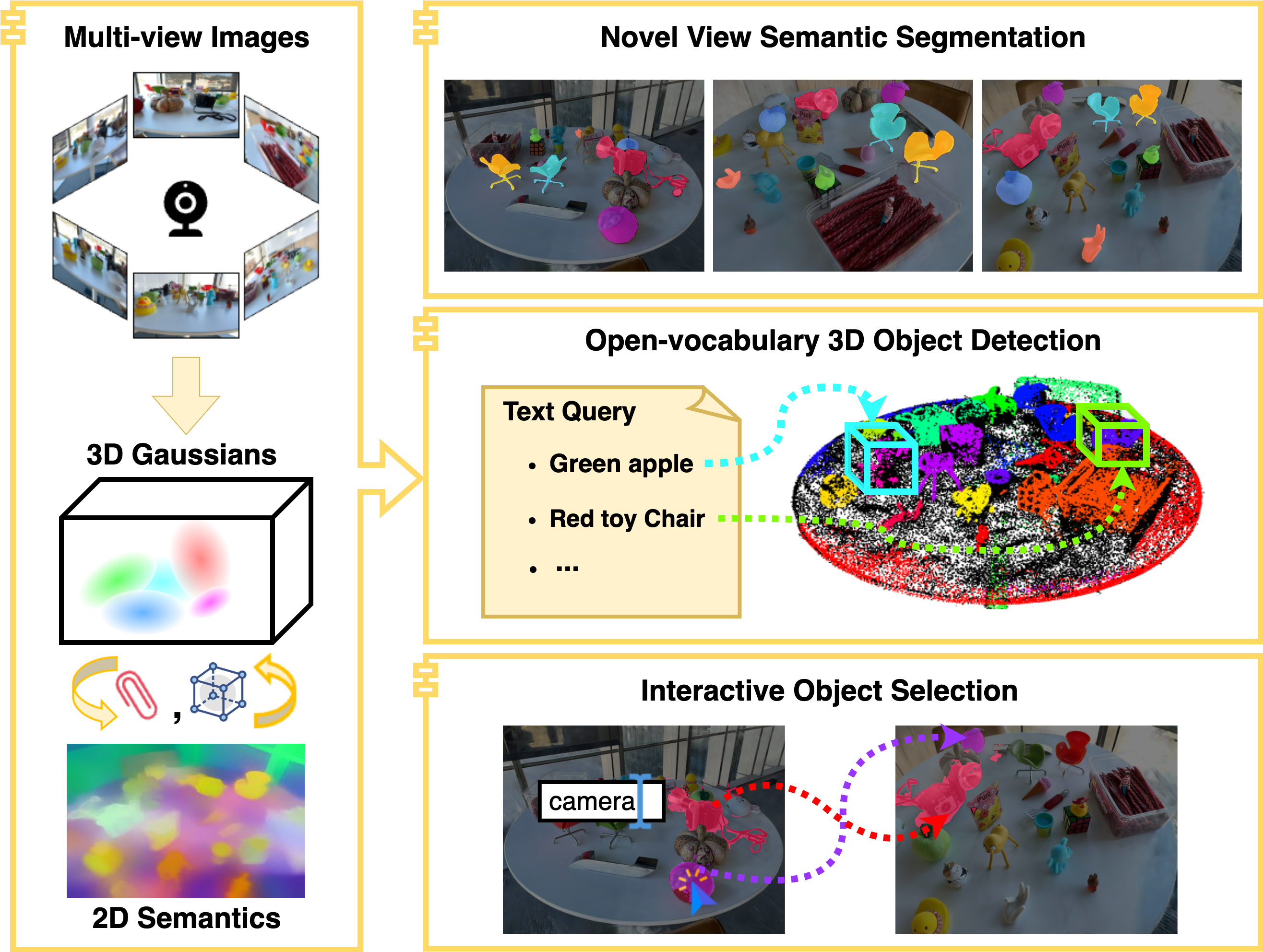}\\
   \caption{We present a novel framework, \textbf{\textit{FreeGS}}, to inject 2D semantics (\includegraphics[height=1.0em]{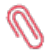}) into 3DGS (\includegraphics[height=1.0em]{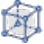}), without the need of any 2D labels. After a 2D-3D collaborative bootstrapping learning strategy, the model can support versatile applications, such as novel-view 2D segmentation, open-vocabulary 3D detection, and interactive object selection.}
   \label{fig:teaser}
\end{figure}

Unlike NeRF, 3DGS utilizes discrete Gaussians to represent scenes and leverages tile-based rasterization for acceleration, enabling real-time rendering at high resolutions. This explicit representation allows 3DGS to perform flexible content manipulations, such as adding, removing, or editing elements within a scene. Recent approaches~\cite{zhou2023feature,qin2023langsplat,ye2023gaussian} have incorporated the aforementioned NeRF-based advancements into 3DGS to enable efficient novel view synthesis and semantic segmentation. To achieve this, in addition to extracting semantic embeddings from CLIP~\cite{CLIP}, 2D class-agnostic masks from the Segment Anything Model (SAM)~\cite{kirillov2023SAM} are also employed as auxiliary constraints on semantic learning to refine segmentation details.

However, over-reliance on 2D supervision imposes certain limitations that hinder 3DGS to represent 3D scenes efficiently and accurately. Firstly, previous methods require complex data preprocessing, particularly the generation of SAM-based pseudo masks \cite{liao2024clip-gs,qin2023langsplat,ye2023gaussian,silva2024contrastive}, which is often cumbersome and time-consuming. Secondly, when processing multi-view images, 2D foundational models \cite{CLIP,li2022language} struggle to maintain consistent object identities and semantics across different views. This lack of view-consistent 2D supervision limits 3DGS to capture accurate semantics across multiple perspectives, thus restricting its accuracy not only in novel-view segmentation but also in tasks like interactive object selection or object perception in 3D space. Recent advances have attempted to address these issues by utilizing either predefined 2D view-consistent identity labels \cite{ye2023gaussian} or 3D self-training loss \cite{silva2024contrastive,liao2024clip-gs}. However, these approaches remain constrained by complex data preparation and are sensitive to the quality of preprocessing outcomes.

To resolve these challenges, we present \textbf{\textit{FreeGS}}, an unsupervised 3DGS framework for view-consistent 3D scene understanding that uses only raw images as input, without requiring any labels. The core of \textbf{\textit{FreeGS}} is the IDentity-coupled Semantic Field (IDSF), which holds both semantic embeddings and view-consistent instance indices for each Gaussian. With only multi-view images as input, we optimize IDSF through a bootstrapping process that alternates between the Union-space 3D Gaussian Clustering phase and the Multi-level 2D Semantic Distillation phase. In the 3D clustering step, semantic information collaborates with the intrinsic properties of Gaussians (e.g., geometry and appearance) to extract unsupervised instances coherently. In the 2D distillation step, the extracted 3D instance indices help to constrain the correspondence of multi-view images to learn view-consistent semantic embeddings. Consequently, semantics and view-consistent instances are optimized complementarily in a self-bootstrapped manner. However, the 3D semantics in IDSF are still indirectly injected from the 2D plane, which may lead to optimization difficulty incurred by the rendering process of 3DGS. To address this, we further introduce a joint contrastive loss to bridge the 2D-3D discrepancy and boost the learning efficacy.

Integrating these components, our method achieves an efficient and compact framework for view-consistent 3D scene understanding. As shown in Fig. \ref{fig:teaser}, our \textbf{\textit{FreeGS}} supports various tasks such as novel-view semantic segmentation, interactive object selection, and 3D object detection. In summary, our contributions are as follows:
\begin{itemize}
    \item We propose \textbf{\textit{FreeGS}}, which leverages the IDentity-coupled Semantic Field to explicitly encode both semantics and instance identities for Gaussians. 
    \item We propose a bootstrapping strategy to effectively optimize the IDentity-coupled Semantic Field without accessing any labels.
    \item Without complex data preprocessing, our \textbf{\textit{FreeGS}} performs comparably to or better than state-of-the-art methods in open-vocabulary 3D scene understanding.
\end{itemize}

\section{Related Work}
\label{sec:relatedwork}

\textbf{3D Gaussian Splatting.}
Recently, 3D scene reconstruction using 3DGS~\cite{kerbl20233DGS} has gained prominence due to its fast rendering capabilities and impressive visual quality. Unlike the implicit radiance field in NeRF~\cite{mildenhall2021nerf}, 3DGS represents the 3D scene with millions of Gaussians, allowing for more flexible access to the radiance field. Building on the success of 3DGS, several subsequent works have integrated it with 2D foundation models for various applications, such as 3D scene generation, editing, and understanding. Among these methods, one popular trend involves adding constraints from 2D space (\textit{e.g.,} Diffusion models) to enable quality enhancement or promptable editing/generation~\cite{tang2023dreamgaussian,yi2023gaussiandreamer,tang2024lgm,yang2024gaussianobject,GaussianEditor}. Meanwhile, some concurrent approaches incorporate extra learnable parameters to the original Gaussians for applications beyond 3D reconstruction. For example, some methods \cite{wu2024sc4d,zeng2024stag4d,ren2023dreamgaussian4d} introduce an extra-temporal dimension into 3DGS for dynamic scene reconstruction, also known as 4D scene synthesis. 

\begin{figure*}[ht]
  \centering
   \includegraphics[width=1\linewidth]{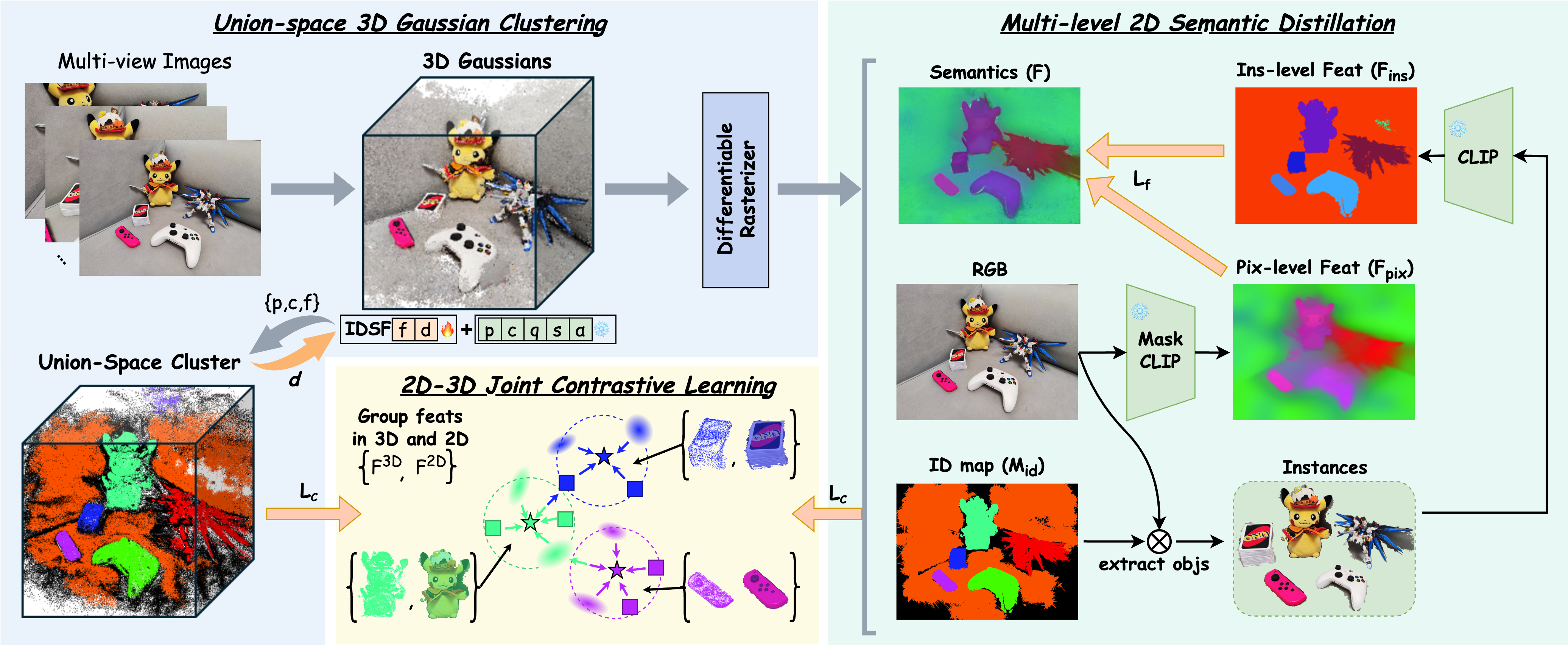}
   \caption{Framework overview of \textit{\textbf{FreeGS}}. The framework consists of three key components: Union-space 3D Gaussian Clustering, Multi-level 2D Semantic Distillation, and 2D-3D Joint Contrastive Learning. In the 3D space, Gaussians equipped with the IDentity-coupled Semantic Fields (IDSF) are input to the union-space clustering module to extract view-consistent instance indices. Subsequently, the IDSF is rendered onto 2D space and supervised by multi-level features from foundational models. Additionally, a 2D-3D joint contrastive loss is applied between instance-aware 3D features and rendered 2D features to enhance the compactness and discrimination of semantics in the joint feature space. The alternating updates of the semantic field and the instance clustering bootstrap view-consistent semantics in Gaussians, without relying on any 2D labels.}
   \label{fig:network}
\end{figure*}

\noindent\textbf{Feature Field Distillation.} The integration of semantic feature field into radiance field is widely explored in NeRF-based approaches~\cite{semantic-nerf,kobayashi2022decomposing,fan2022nerf}. Early attempts like Semantic NeRF \cite{kobayashi2022decomposing} successfully incorporate geometry, appearance, and semantics into the radiance field, showing that precise 3D segmentation can be learned from sparse and noisy 2D annotations. Later, DFF~\cite{kobayashi2022decomposing}, LERF \cite{kerr2023lerf}, and 3DOVS \cite{liu2023weakly} distill semantic feature from CLIP \cite{CLIP} or DINO \cite{DINO} into NeRF to achieve open-world 3D perception. Recently, these advances from NeRF have been introduced to 3DGS \cite{lan20232d, hu2024semantic, huang2023point} to form a more efficient and scalable 3D reconstruction \& understanding framework. Beyond CLIP, the Segment Anything Model (SAM) \cite{kirillov2023SAM} is commonly used to generate 2D masks for training views, serving as a weak supervision to improve mask quality. 
LangSplat \cite{qin2023langsplat} uses CLIP to supervise the newly-added semantic embeddings and adopt multi-granularity 2D segmentation masks from SAM for fine-grained perception. However, the learned semantic embedding fails to capture view-consistent identity without an effective cross-view constraint. Gaussian Grouping \cite{ye2023gaussian} assigns each Gaussian a view-consistent ID for cross-view perception. However, this method requires supervision from predefined view-consistent labels, which are generated through complex preprocessing with SAM and tracking models \cite{cheng2023tracking}. 
Some concurrent works also attempt to enhance the feature field with various motivations. For example, N2F2 \cite{bhalgat2024n2f2} improves the feature field to be scale-aware and fairly reduces the consuming time of training and testing. CGC \cite{silva2024contrastive} and OpenGaussian \cite{wu2024opengaussian} both design contrastive losses to avoid the need for view-consistent 2D labels. However, all of them still need frame-wise SAM masks to group the Gaussians of the same identity.
Building upon a 3DGS-based framework, we introduce a bootstrapping 3D clustering strategy to learn view-consistent semantic fields. This algorithm not only eliminates the time-consuming SAM-based 2D mask generation process but also contributes to prompt-unified segmentation in both 2D and 3D scenarios. 

\section{Methodology}
\label{sec:method}
In this paper, we propose \textbf{\textit{FreeGS}} for achieving view-consistent 3D scene understanding without the need for 2D/3D labels. The overall framework of \textbf{\textit{FreeGS}} is illustrated in Fig.~\ref{fig:network}. Given only multi-view images as input, the IDentity-coupled Semantic Field (IDSF) is designed to capture both semantic embeddings and view-consistent instance indices. The IDSF is optimized alternately between two stages: Union-space 3D Gaussian Clustering and Multi-level 2D Semantic Distillation. In 3D space, the Union-Space Clustering is used to extract the view-consistent instance indices. Then, the produced indices are projected onto 2D space to regularize the semantic injection with foundational models \cite{CLIP,dong2023maskclip}. Additionally, 2D-3D Joint Contrastive Learning is employed to aid in the complementary of 3D geometry and 2D semantics during this bootstrapping process.   

\subsection{Preliminaries on 3D Gaussian Splatting}
3D Gaussian Splatting (3DGS) \cite{kerbl20233DGS} represents a 3D scene with millions of Gaussians, where each Gaussian is parameterized with a 3D position vector $\mathbf{p}=\{x, y, z\}\in \mathbb{R}^3$, a color vector $\mathbf{c}\in \mathbb{R}^3$, a rotation quaternion $\mathbf{q} \in \mathbb{R}^4$, a scaling vector $\mathbf{s} \in \mathbb{R}^3$ and an opacity value $\alpha\in \mathbb{R}$, denoted as $\Theta_i=\{\mathbf{p}_i, \mathbf{c}_i, \mathbf{q}_i, \mathbf{s}_i, \alpha_i\}$. To reconstruct the 3D scene from multi-view images with camera parameters $\sigma$, a tile-based differential rendering strategy is adopted to project the 3D Gaussians onto 2D planes according to $\sigma$. The pixel color $C$ is computed by $\alpha$-blending the Gaussians in a front-to-back depth manner \cite{kopanas2021point}:
\begin{equation}
{C} = \sum_{i \in \mathcal{N}} T_i \alpha'_i \mathbf{c}_i,
\label{eq:3DGSplane}
\end{equation}
where $\mathcal{N}$ denotes the number of Gaussians in the tile and $\alpha'_i$ represents the Gaussian's final influence on certain pixels. The transmittance $T_i$ is calculated by $T_i = \prod_{j=1}^{i-1} (1 - \alpha'_j)$. 
During training, the Gaussians are initialized with the sparse points from SfM \cite{snavely2006photo}. Then, they are optimized by constructing a reconstruction loss between ground truth image $I_{GT}$ and the reconstructed image $I$ in various views:
 \begin{equation}
     \mathcal{L}_{R} = (1-\lambda)\mathcal{L}_1({I},I_{GT}) + \lambda \mathcal{L}_{SSIM} (I,I_{GT}). 
 \label{eq:L_render}
\end{equation}
Here, $\lambda=0.2$ and the pixels of reconstructed image $I$ are obtained by Eq. \ref{eq:3DGSplane}. The explicit 3D representations with Gaussians and the tile-based rasterizer collaborate to improve the rendering efficiency and enable flexible scene editing via directly manipulating the desired Gaussians. 

\subsection{Union-space 3D Gaussian Clustering}
Previous methods \cite{qin2023langsplat,silva2024contrastive,liao2024clip-gs} usually incorporate additional regularizations during the 2D feature distillation stage to enhance consistent semantics across views. However, due to the lack of 3D environmental perception in 2D supervision, these methods still face challenges in effectively addressing various 3D scene understanding tasks. Compared to 2D semantics, 3D Gaussians can offer additional low-level priors on appearance and geometry for each instance. This inspires us to incorporate the 3D correspondence of Gaussians with 2D semantics to optimize view-consistent semantics. To this end, we propose a union-space Gaussian Clustering strategy to group Gaussians and extract their identity indices that are consistent across views.

\noindent\textbf{IDentity-coupled Semantic Field.}
A robust semantic field is crucial for 3DGS to achieve consistent and view-insensitive segmentation. 
Previous methods \cite{silva2024contrastive,qin2023langsplat} usually attach a high-dimensional semantic vector to each Gaussian and employ self-supervision strategies to ensure consistent vector distribution across views. However, this design often leads to object mismatch when switching views. To address this issue, we propose the IDentity-coupled Semantic Field to explicitly capture the semantic representation and cross-view instance index for each Gaussian. Specifically, the IDSF can be represented as $\{\mathbf{f}_i, {d}_i\}$, where $\mathbf{f}_i \in \mathbb{R}^D$ is the view-independent semantic vector and ${d}_i \in \mathbb{R}$ is the cross-view instance index. These elements will serve as the basic units to perform alternating updating between 3D and 2D spaces.  

\noindent\textbf{Union-space Clustering.} 
Based on IDSF, we perform clustering on geometry-appearance-semantics union space to extract the index $\mathbf{d}_i$ for each Gaussian. To group Gaussians of the same objects in 3D space, we define the union clustering space as $\mathcal{S}_{\text{union}} = \mathcal{S}_{\text{pos}} \cup \mathcal{S}_{\text{app}} \cup \mathcal{S}_{\text{sem}}$. In this context, $\mathcal{S}_{\text{pos}}$ is composed of the Gaussian's center coordinate $\mathbf{p}$, $\mathcal{S}_{\text{app}}$ is the gather of view-independent color part $\mathbf{c}'$, and $\mathcal{S}_{\text{sem}}$ encompasses the semantic feature $\mathbf{f}'$ after PCA dimensionality reduction. Specifically, we first extract the semantic-injected points $X_{\Theta_i} = \{\mathbf{p}_i, \mathbf{c}'_i, \mathbf{f}'_i\} \in \mathbb{R}^{\{3+3+6\}}$ from 3DGS. Then, we apply normalization to each sub-space and concatenate them in the channel dimension. Finally, we send them into HDBSCAN~\cite{mcinnes2017hdbscan} to perform clustering in the union space $S_{union}$ and obtain a set of clustered groups $\{\mathcal{G}_i\}_{i=1}^{N_g}$. Here, $N_g$ indicates the number of groups. Each group consists of Gaussians sharing similar appearance, geometry and semantics, suggesting they belong to the same object. We assign the group index to each Gaussian to form the view-consistent instance index $\mathbf{d}_i$, which will be used to aid in 2D semantic distillation. 

\noindent\textbf{Feature Field Smoothing.} Although the multi-level priors from Gaussians are involved in the clustering space, it may still be challenging to successfully group all the Gaussians belonging to objects with complex structure. To mitigate the impact of noisy points during clustering, we introduce a 3D smoothing loss for IDSF to improve the feature field continuity from a local perspective. Practically, we organize the neighbor tree of 3D Gaussians and sample $T$ Gaussians to perform feature smoothing with their $K$-nearest neighbors:  
\begin{equation}
    \mathcal{L}_{S} = \frac{1}{TK} \sum_{i=1}^{T} \sum_{j=1}^{K} \left(1 - \text{sim}\left(\mathbf{f}_i, \mathbf{f}_j\right)\right), 
\label{eq:smoothloss}
\end{equation}
where sim$(\cdot)$ denotes cosine similarity. By constraining the semantics of adjacent Gaussians, the proposed smoothing loss can effectively improve the completeness of clustered groups. Besides, it can help to reduce the grid artifacts caused by feature downsampling in 2D semantic distillation.    

\subsection{Multi-level 2D Semantic Distillation}
To infuse 3D Gaussians with semantics, we utilize the widely adopted semantic distillation technique with CLIP models. Different from previous methods that rely on pixel-wise feature constraints, we introduce a multi-level 2D semantic distillation strategy to enhance the semantic injection from both local and global perspectives. 
Similar to Eq.~\ref{eq:3DGSplane}, we $\alpha$-blend the semantic feature $\mathbf{f}_i$ onto 2D plane by:
\begin{equation}
\mathbf{F} = \text{MLP}\left(\sum_{i \in \mathcal{N}} T_i \alpha'_i \mathbf{f}_i\right), 
\end{equation}
where $\text{MLP}(\cdot)$ indicates we use an MLP layer to achieve dimension alignment between the Gaussian semantic feature and the CLIP feature. $\mathbf{F}\in\mathbb{R}^{512}$ represents the rendered feature map with post-process. 

\noindent\textbf{Pixel-level Feature Distillation.} 
The large resolution discrepancy between the rendered feature map (\textit{e.g.,} $1080 \times 1440$) and CLIP features (\textit{e.g.,} $14 \times 14$) would lead to detail degradation (also known as grid artifacts). Therefore, we propose pixel-level feature distillation to enhance the details of IDSF. To this end, we exploit MaskCLIP \cite{dong2023maskclip} as a feature extractor to extract 2D spatially vision-text-aligned features from the origin CLIP visual encoder, for pixel-level feature distillation. Specifically, we feed the rendered image $I$ into MaskCLIP and adopt a recent feature super-resolution work named FeatUP \cite{fu2024featup} to increase the feature resolution to $224 \times 224$, which is represented as $\mathbf{F}_{pix}$. To rectify the resolution misalignment between $\mathbf{F}$ and $\mathbf{F}_{pix}$, we also adopt a convolutional layer to downsample $\mathbf{F}$ to the same resolution as $\mathbf{F}_{pix}$, which is denoted as $\hat{\mathbf{F}}$.

\noindent\textbf{Instance-level Feature Distillation.} 
Except for the pixel-level distillation, we introduce an instance-level feature distillation strategy to boost the semantic consistency inside each instance. To achieve this, we utilize the view-consistent instance index $d_i$ obtained at the 3D clustering stage. Concretely, we convert the index $d_i$ into a one-hot vector ${d}'_i$ and project it onto 2D space to produce the ID map $M_{id}$ with an argmax operation:  
\begin{equation}
{M_{id}} = \arg\max \left(\sum_{i \in \mathcal{N}} T_i \alpha'_i {d}'_i \right). 
\label{eq:2dindex}
\end{equation}

The rendered result for each pixel is the weighted summation of $\mathbf{d}'_i$ and, therefore, represents the most likely instance indice after the argmax operation. With $M_{id}$, we can extract the pseudo mask for each detected instance, denoted as $M = \{M_i \in \mathbb{R}^{H \times W} | i = 1,2,...,N_g\}$, and these masks will be processed by DenseCRF for boundary refinement. Subsequently, we apply these class-agnostic masks to the rendered image to extract image patches, which are then input into the CLIP visual encoder. That is to say, we obtain the instance-level feature map by:
\begin{equation}
    \mathbf{F}_{ins} = \sum_{i=1}^{N_g} {M}_i \odot \text{CLIP}\left({M}_i \odot I\right),
\end{equation}
where $\odot$ is element-wise multiplication. Finally, we construct the multi-level feature distillation loss as:
\begin{equation}
    \mathcal{L}_F = \|\mathbf{F}_{pix} - \hat{\mathbf{F}}\|_1 + \gamma  \|\mathbf{F}_{ins} - \mathbf{F}\|_1,
\label{eq:2dfeatloss}
\end{equation}
where $\gamma$ is a trade-off to balance the impact of two items.

\subsection{Bootstraping via 2D-3D Joint Contrastive Learning}
The two stages mentioned above allow the semantic field $\mathbf{f}_i$ and instance index $d_i$ to update alternately. As training progresses, well-aligned semantic fields will facilitate the extraction of more accurate instance indices. This will also provide more precise semantic guidance to supervise the semantic field in turn. This bootstrapping strategy enables the successful incorporation of view-consistent semantics into Gaussians without the need to access any labels. 

However, without explicit supervision to constrain the view consistency, clustering in 3D space would easily incur noisy grouping results in cluttered scenarios. Therefore, we propose a 2D-3D joint contractive learning strategy to compact the features within the same object and separate those from different objects.   
In detail, given a clustered group $\mathcal{G}_i$, we first collect the semantic features of component Gaussians to form the 3D feature group $\mathcal{F}_i^{3D}=\{\mathbf{f}_{j}\mid d_j = i\}$, and the features of all 2D pixels within the corresponding group mask $M_i$ to form the 2D feature group $\mathcal{F}_i^{2D}=\{\mathbf{f}_{v}\mid M_i\left(v\right) = 1\}$. Here, $v$ represents a certain pixel. Then, we combine $\mathcal{F}_i^{3D}$ and $\mathcal{F}_i^{2D}$ into a 2D-3D joint feature group, denoted as $\mathcal{F}^{\mathcal{G}_i}=\mathcal{F}_i^{3D} \cup \mathcal{F}_i^{2D}$. 

Meanwhile, to facilitate re-merging of instance fragments in the next clustering step, for each feature group $\mathcal{F}^{\mathcal{G}_i}$, we set groups that exceed the feature similarity threshold as positive samples and collect their mean features into a set $\bar{\mathcal{F}}^{\mathcal{G}_{i+}}$, while the remaining are collected into $\bar{\mathcal{F}}^{\mathcal{G}_{i-}}$ in the same way. The 2D-3D joint contrastive loss $\mathcal{L}_C = \frac{1}{N_g}\sum_{i=1}^{N_g}\mathcal{L}_C^{\left(i\right)}$, where each $\mathcal{L}_C^{\left(i\right)}$ is defined as:
\begin{equation}
\label{eq:3dc}
    \mathcal{L}_C^{\left(i\right)} = - \sum_{j=1}^{|\mathcal{F}^{\mathcal{G}_i}|} \log{\frac{\sum_{\mathcal{G}_{i+}}  \exp{\left(\text{sim}\left(\mathcal{F}_j^{\mathcal{G}_i},\mathcal{\bar{F}}^{\mathcal{G}_{i+}} \right) \mathbin{/} \tau \right)}}  {\sum_{\mathcal{G}_{i-}} \exp{\left(\text{sim}\left(\mathcal{F}_j^{\mathcal{G}_i},\mathcal{\bar{F}}^{\mathcal{G}_{i-}} \right) \mathbin{/} \tau \right)}}},
\end{equation}
where $\tau$ denotes a temperature parameter and is set to 0.1 in all experiments.

\subsection{Training and Testing}
\textbf{Training.} The entire training process includes two phases. In the first phase, the 3DGS is optimized with Eq.~\ref{eq:L_render} for scene reconstruction. The Gaussians that greatly capture scene appearance and geometry can provide reliable priors to improve the stability of union-space clustering. Then, in the second phase, we incorporate the IDSF into Gasussians and optimize them with the loss function:
\begin{equation}
    \mathcal{L} = \mathcal{L}_{F} + \lambda_{C}\mathcal{L}_{C} + \lambda_{S}\mathcal{L}_{S}.
\end{equation}
Here, $\mathcal{L}_{F}$, $\mathcal{L}_{C}$ and $\mathcal{L}_{S}$ are multi-level feature distillation loss (Eq.~\ref{eq:2dfeatloss}), 2D-3D contrastive loss (Eq.~\ref{eq:3dc}) and smoothing loss (Eq.~\ref{eq:smoothloss}). $\lambda_{C}$ and $\lambda_{S}$ are two trade-offs.

\noindent\textbf{Testing.} By incorporating view-consistent semantics into 3DGS, our FreeGS can uniformly address a variety of 3D scene understanding tasks, such as open-vocabulary novel-view semantic segmentation, 3D object detection, and interactive object selection. More importantly, the proposed IDSF successfully replaces the previous view-dependent inference mode in 2D plane with a view-independent inference pipeline in 3D space. 

Specifically, we abandon the previous process of querying separately under each view, and instead propose a match-by-propose approach as an advanced replacement. 
Once trained, we can easily extract 3D instance proposals (\textit{i.e.,} clustered groups) and acquire their average features and 3D bounding boxes from the component Gaussians. For each query, we compute relevancy scores between the query embedding and the average feature of each group following LERF \cite{kerr2023lerf}, and choose the group with the highest score. Then, based on the acquired 3D proposals, we can easily generate the corresponding 2D segmentation map for novel-view segmentation and 3D box for 3D object detection. For interactive prompts like click, we are able to directly access the Gaussian according to the click coordinate in 3D space and return the corresponding instance index to produce 2D masks or 3D boxes. 

\begin{figure*}[t]
  \centering
   \includegraphics[width=1.0\linewidth]{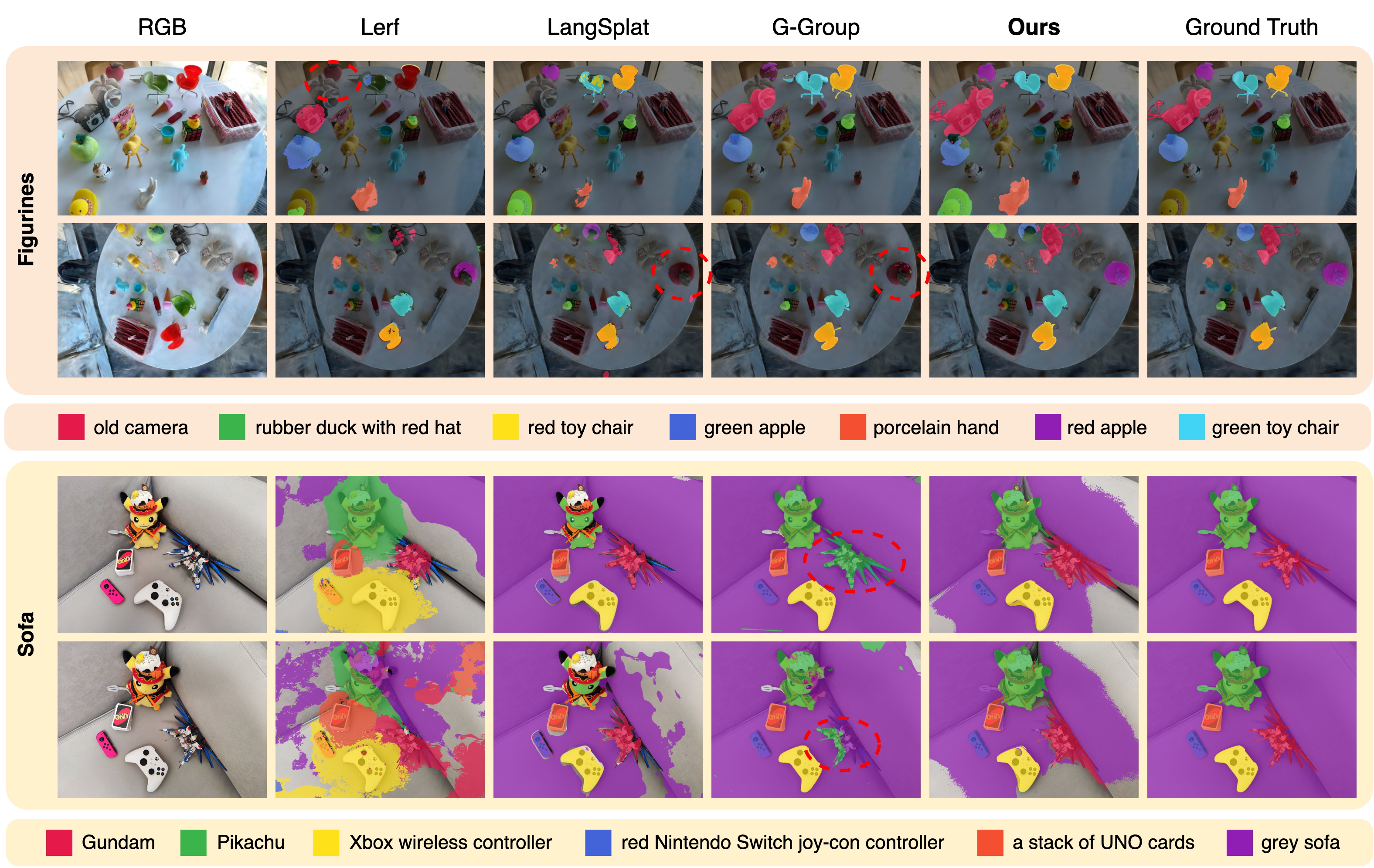}
   \caption{Qualitative comparisons of different methods on LERF-Mask and 3D-OVS dataset. Our method successfully segment instances with consistency across different views.}
   \label{fig:2dseg}
   \vspace{5pt}
\end{figure*}

\section{Experiments}
\subsection{Datasets and Metrics}
\textbf{Datasets.} To evaluate the effectiveness of our method, we conduct experiments on two widely used datasets: LERF-Mask \cite{ye2023gaussian} and 3D-OVS \cite{liu2023weakly}. 
LERF-Mask dataset consists of three scenes with dense annotations. These scenes are categorized as ``long-tail scenes" of LERF-Localization dataset \cite{kerr2023lerf}, which presents great challenges for scene understanding tasks, especially in unsupervised settings. 3D-OVS dataset contains uncommon objects like ``Gundam" and ``Pikachu", posing even greater challenges to the task. We select five scenes following the criteria of ~\cite{qin2023langsplat}. Except for novel-view perception, we evaluate the proposed method on 3D object detection and establish a benchmark on ScanNet \cite{dai2017scannet}, which is a widely used dataset with complex indoor scenes for 3D scene understanding. In detail, we conduct experiments on three scenes in ScanNet: scene0060, scene0381, and scene0462. For each scene, we sample 10\% views for reconstruction, with the sampling strategy based on image sharpness to avoid motion blur.

\noindent\textbf{Metrics.} For 2D open-vocabulary segmentation task on LERF-Mask and 3D-OVS datasets, we use the mIoU metric to measure segmentation accuracy.
For 3D open-vocabulary detection task on ScanNet dataset, we observe that the provided 2D annotations are extremely coarse. So we use 3D box mIoU and recall to evaluate the performance. The IoU threshold is set to 0.25 and 0.5 for calculating the recall.

\begin{table}[t]
  \centering
  \small
  \renewcommand\arraystretch{1.25}
  \begin{tabular}{p{1.2cm}<{\centering}|p{0.4cm}<{\centering}p{0.4cm}<{\centering}|p{0.7cm}<{\centering}p{1.0cm}<{\centering}p{0.85cm}<{\centering}p{0.85cm}<{\centering}}
    \Xhline{1pt}
    {Method}&SAM&{PP}&{Ramen}&{Figurines}&{Teatime}&{Average}\\
    \hline
    LERF &\ding{55}&\ding{55}&28.3&33.5&49.7&37.2\\
    G-Group &\checkmark&\checkmark&77.0&69.7&71.7&72.8\\
    LangSplat &\checkmark&\checkmark&69.7&54.7&65.4&63.3\\
    \hline
    Ours&\ding{55}&\ding{55}&77.5&62.6&68.5&69.5\\
    \Xhline{1pt}
  \end{tabular}
  \caption{Quantitative comparisons of 2D novel view segmentation on LERF-Mask dataset in terms of mIoU. The item ``SAM'' denotes whether 2D masks from SAM are used as supervision during training. The ``PP'' indicates whether an additional preprocessing is conducted during training.}
  \label{tab:lerf-mask}
\end{table}

\begin{table}[t]
  \centering
  \small
  \renewcommand\arraystretch{1.25}
  \begin{tabular}{p{1.2cm}<{\centering}p{0.66cm}<{\centering}p{0.7cm}<{\centering}p{0.66cm}<{\centering}p{0.7cm}<{\centering}p{0.7cm}<{\centering}p{0.8cm}<{\centering}}
    \Xhline{1pt}
    {Method}&{Bed}&{Bench}&{Room}&{Sofa}&{Lawn}&{Average}\\
    \hline
    LERF&73.5&53.2&46.6&27.0&73.7&54.8\\
    G-Group&64.9&74.0&79.0&70.2&96.8&77.0\\
    LangSplat&79.7&82.8&87.5&77.1&94.1&84.2\\
    \hline
    Ours&79.5&86.1&70.1&81.9&67.4&77.0\\
    \Xhline{1pt}
  \end{tabular}
  \caption{Quantitative comparisons of novel view semantic segmentation on 3D-OVS dataset in terms of mIoU.}
  \label{tab:3dovs}
\end{table}

\begin{figure*}[t]
  \centering
   \includegraphics[width=1\linewidth]{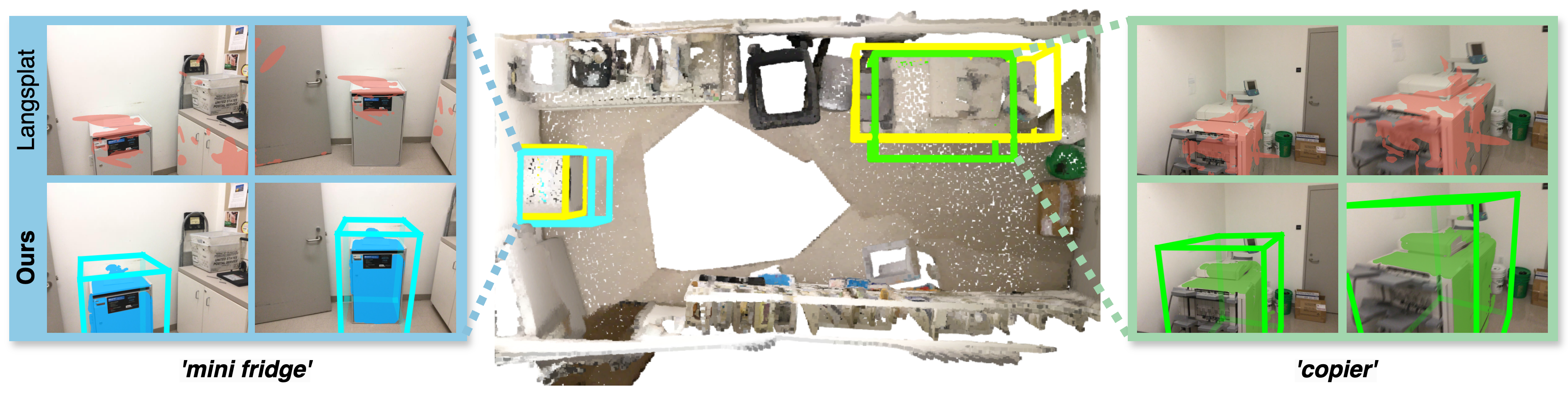}
   \caption{Qualitative comparisons of different methods on ``Scene0462'' of the ScanNet dataset. We use yellow lines to draw the ground truth 3D bounding boxes. For the two queries in the figure, LangSplat generates bounding boxes as large as the whole scene because of the existence of noise points, so we do not show box visualization for LangSplat. While our method demonstrates much more accurate localization and segmentation.}
   \label{fig:3dseg}
   \vspace{5pt}
\end{figure*}

\begin{table*}[!ht]
\centering
\renewcommand\arraystretch{1.25}
\begin{tabular}{c|ccc|ccc|ccc}
\Xhline{1pt}
   \multirow{2}{*}{Model} & \multicolumn{3}{c|}{Scene0060} & \multicolumn{3}{c|}{Scene0381} & \multicolumn{3}{c}{Scene0462}  \\ 
  & mIoU & Recall50 & Recall25 & mIoU & Recall50 & Recall25 & mIoU & Recall50 & Recall25    \\ 
  \hline
Langsplat &5.6&0.0&8.3&13.9&7.1&14.3&3.3&0.0&0.0  \\   
Ours &13.7&8.3&25.0&26.2&21.4&42.9&14.5&11.5&26.9\\ 
\Xhline{1pt}
\end{tabular}
\caption{Quantitative comparisons of open vocabulary 3D detection on the ScanNet dataset. We report the 3D box's mIOU, Recall50 and Recall25. }
\label{tab:scannet}
\end{table*}

\subsection{Implementation Details}
We implement the proposed method based on 3DGS ~\cite{kerbl20233DGS} and conduct all experiments on an Nvidia A800 GPU. We use the cuML library \cite{raschka2020machine} to support all machine learning algorithms in our work, such as HDBSCAN and PCA, for GPU acceleration. For scene reconstruction, we follow Mini-Splatting \cite{fang2024mini} for a more uniform spatial distribution of 3D Gaussians. For each scene, the first training phase takes 30k iterations and the second phase for 7k iterations.
We set the hyperparameters of $\lambda_{S}$ and $\lambda_{C}$ as 0.1 and 0.05 respectively, and set the feature vector dimension $D = 128$. We use the Adam optimizer with a learning rate of 0.0025 and 0.0001 for GS and CNN layers respectively. For $L_F$, we set $\gamma = 0.3$. For $L_S$, we set $K = 5$ and $T$ as 0.1\% of the total number of Gaussians. We apply $L_S$ and $L_C$ every 10 iterations for acceleration. 
For hyperparameters used in HDBSCAN, we set $min\_samples = 20$ in experiments on LERF-Mask, and $min\_samples = 60$ in experiments on 3D-OVS and ScanNet.

\subsection{Comparisons with State-of-the-art Methods}
We compare our method with state-of-the-art public approaches, including the NERF-based LERF \cite{kerr2023lerf}, as well as the 3DGS-based Gaussian Grouping (G-Group) \cite{ye2023gaussian} and LangSplat \cite{qin2023langsplat}. \\
\noindent\textbf{Novel View 2D Segmentation.}
Tab. \ref{tab:lerf-mask} illustrates the quantitative comparisons on the LERF-Mask dataset. As we can see, our method not only outperforms the unsupervised method LERF by a large margin but also exhibits comparable capacity to those SAM-supervised methods. In the ``ramen" scene, we even achieve the best metric among all compared methods. 

Tab.~\ref{tab:3dovs} reports the quantitative results on the 3D-OVS dataset. We found that previous methods did not split datasets during training, so we retrained the methods with only training views and tested them on testing views to evaluate generalization. Note that LERF fails to reconstruct some scenes in this setting, so we use the metric reported in LangSplat. As shown, without any preprocessing, our method achieves comparable overall performance to Gaussian Grouping. Among the five scenes, our method achieves the best performance on ``bench" and ``sofa" scenes, with a significant improvement of 3.99\% and 6.3\% in mIoU metric, respectively. 

Fig.~\ref{fig:2dseg} illustrates the qualitative comparison results. As shown, our method is effective at improving the segmentation accuracy and consistency across views. In particular, for hard cases like ``red apple" in the ``figurines" scene, we observe that LangSplat and Gaussian Grouping both fail to segment correctly under drastically changed views. In contrast, our method produces view-consistent segmentation results successfully, thanks to the IDSF design and view-independent inference pipeline.\\
\begin{table}[!t]
  \centering
  \renewcommand\arraystretch{1.25}
  \begin{tabular}{p{1.6cm}<{\centering}|p{1.8cm}<{\centering}p{1.7cm}<{\centering}p{1.6cm}<{\centering}}
    \Xhline{1pt}
    {Method}&{Preprocess(m)}&{Training(m)}&{Total(m)}\\
    \hline
    LERF&0&54&54\\
    G-Group&24&68&92\\
    LangSplat&141&105&246\\
    \hline
    Ours&0&62&62\\
    \Xhline{1pt}
  \end{tabular}
  \caption{Efficiency comparisons of different methods on the ``ramen'' scene of the LERF-Mask dataset. Our method significantly surpasses previous SAM-based approaches in efficiency.}
  \label{tab:time}
\end{table}\\
\noindent\textbf{Open Vocabulary 3D Detection.}
Previous methods fail to encode semantic field and view-consistent identity indices simultaneously in 3DGS or NeRF, hindering their application to 3D perception tasks. To better evaluate the performance on the 3D detection task, we extend LangSplat to perform 3D detection by generating 3D bounding boxes for Gaussians with high relevancy scores to text queries. 
Tab.~\ref{tab:scannet} illustrates the quantitative comparison results. As shown, without a specific design for the task, our method outperforms LangSplat by a large margin, which further demonstrates the superiority of our method on the 3d scene understanding task. Besides, Fig.~\ref{fig:3dseg} shows the visual comparison results. We observe that Langsplat suffers from severe noisy points and cross-view inconsistency, which leads to imprecise box predictions (often as large as the whole scene). In contrast, our clustering-based method builds a more compact and discriminative feature field, successfully producing accurate masks and boxes for each text query.\\
\textbf{Preprocessing \& Training Time.} 
Tab.~\ref{tab:time} illustrates the efficiency comparisons on the ``ramen'' scene of the LERF-Mask dataset. Although LERF avoids the preprocessing phase and achieves the fastest, its accuracy is notably low as detailed in Tab.~\ref{tab:lerf-mask}. For SAM-based methods G-Group and LangSplat, preprocessing times are 24 and 141 minutes respectively, accounting for over 35\% and 134\% of their training time. The complex and time-consuming preprocessing significantly impairs the models' efficiency and practical applicability. 

\subsection{Ablation Studies}
\begin{figure}[t]
  \centering
   \includegraphics[width=1\linewidth]{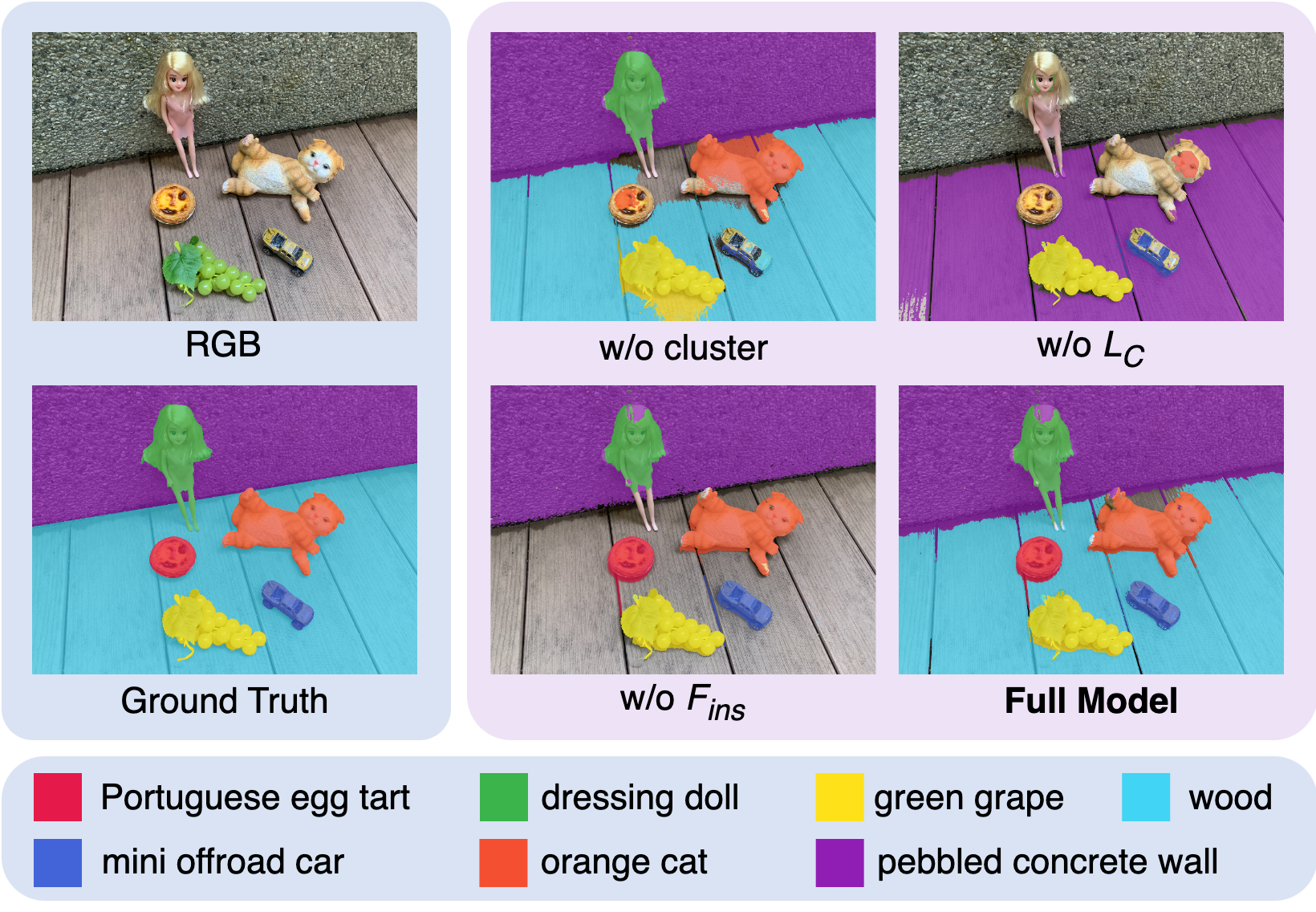}
   \caption{Visualization of ablation results on the ``bench" scene of 3D-OVS dataset.}
   \label{fig:ablation}
   \vspace{5pt}
\end{figure}

\begin{table}[t]
\centering
\small
\renewcommand\arraystretch{1.25}
\begin{tabular}{p{1.5cm}<{\centering}|p{1.0cm}<{\centering}p{0.7cm}<{\centering}|p{0.6cm}<{\centering}p{1.0cm}<{\centering}p{1.0cm}<{\centering}}
\Xhline{1pt}
\multirow{2}{*}{Setting}& figurines & bench  & \multicolumn{3}{c}{Scene0060}\\
 & mIoU & mIoU & mIoU & Recall50 & Recall25\\
\hline
w/o cluster&41.5&60.2&N/A&N/A&N/A\\
w/o $L_C$&33.0&34.3&4.9&0.0&12.5\\
w/o $\mathbf{F}_{ins}$&33.3&73.6&11.7&4.2&20.8\\
\textbf{Full Model}&62.6&86.1&13.7&8.3&25.0\\
\Xhline{1pt}
\end{tabular}
\caption{Ablation results. We pick one scene per dataset: the ``figurines" scene of LERF-Mask, the ``bench" scene of 3D-OVS and ``scene0060" of ScanNet dataset.}
\label{tab:ablation}
\end{table}

We conduct comprehensive ablation studies to analyze the impact of each component in our method. The results are presented in Tab. \ref{tab:ablation} and Fig. \ref{fig:ablation}.
Without clustering, our model can only be queried in 2D space and therefore do not support 3D tasks on the ScanNet dataset. Although the performance seems not bad at first sight in this setting, the ``Portuguese egg tart" and ``mini offroad car" cannot be recognized at all.
The 2D-3D joint contrastive learning is primarily designed to stabilize the clustering and pull instance fragments together. Without $\mathcal{L}_{C}$, the clustering results can sometimes be quite different as training proceeds, and instances are often partitioned into fragments. 
During experiments on the ``figurines" scene, we found an interesting observation that MaskCLIP fails to distinguish between ``red apple'' and ``green apple'' while CLIP can do that successfully, and this is in fact our motivation for the design of $\mathbf{F}_{ins}$. Also, as shown in the figure, $\mathbf{F}_{ins}$ helps our method to recognize ``wood", which indicates that $\mathbf{F}_{ins}$ improves our method's instance discrimination capability.

In summary, all these proposed components work in synergy, effectively and positively contributing to excellent 3D scene understanding performance.

\begin{figure}[t]
  \centering
   \includegraphics[width=1\linewidth]{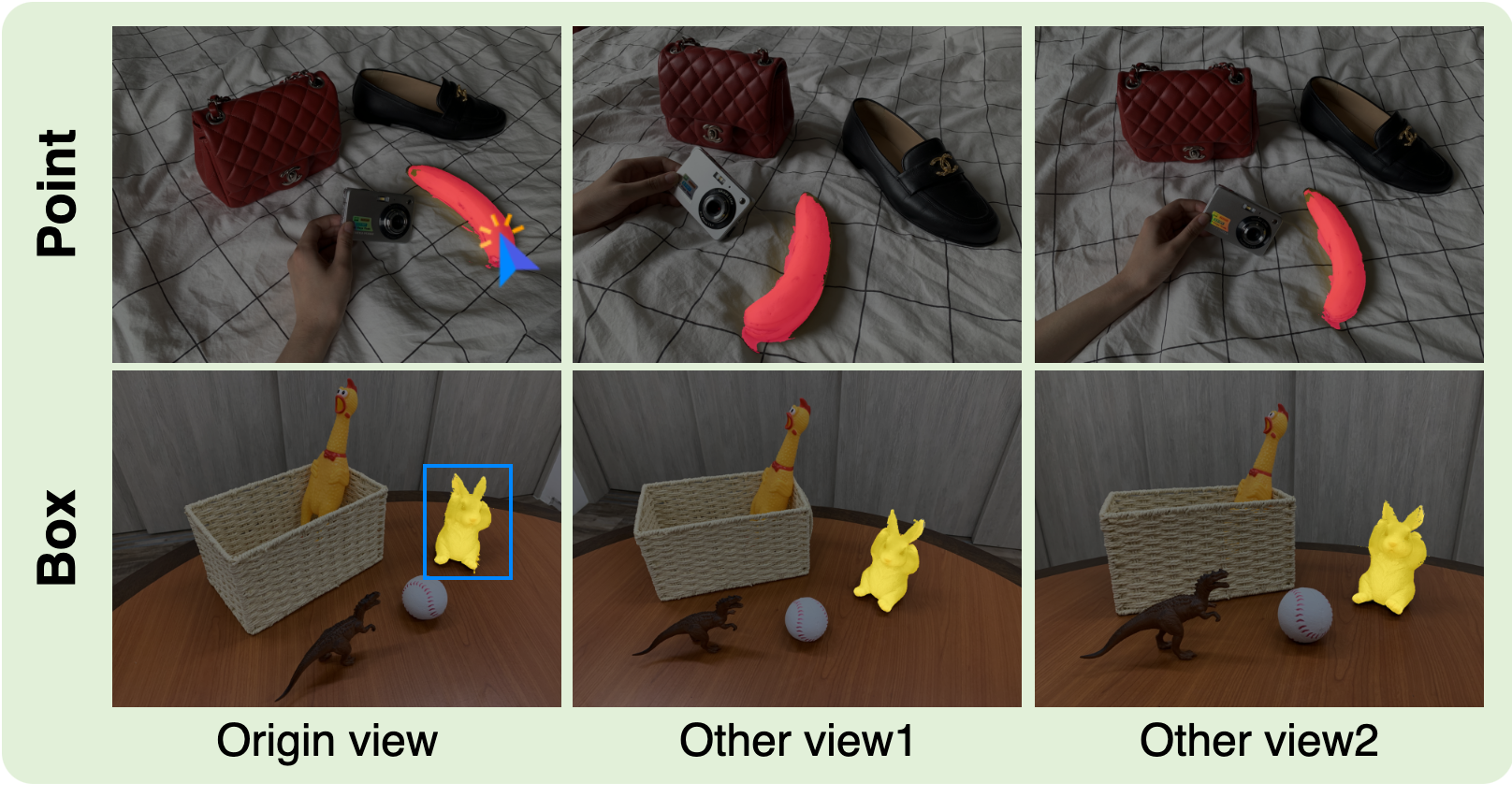}
   \caption{Visualization of interactive object selection results with point or box as prompts. Scenes are ``bed" and ``room" of the 3D-OVS dataset.}
   \label{fig:selection}
\end{figure}

\section{Conclusion}
In this paper, we present \textbf{\textit{FreeGS}}, a novel unsupervised method for 3D scene understanding. 
Based on the proposed IDentity-coupled Semantic Field (IDSF) and 2D-3D collaborative bootstrapping optimization strategy, the proposed method eliminates the need for time-consuming preprocessing and uniformly addresses a variety of tasks, including novel view 2D segmentation, open vocabulary 3D detection and interactive object selection (as shown in Fig. \ref{fig:selection}). 

However, some limitations remain in this label-free training paradigm. First, the introduced semantic fields bring extra memory and computational overhead to 3DGS, slightly impacting the model's efficiency. Second, the Gaussian misalignment phenomenon and the large-scale low-texture background jointly result in suboptimal background segmentation. Furthermore, there is still significant room for improvement in 3D detection performance and we believe that 3D tasks may represent the future direction for 3DGS development.
\section{Acknowledgments}
This work was supported by National Natural Science Foundation of China under Grant 62206039, 62406053, 62293542, 62476048, and the Fundamental Research Funds for the Central Universities (DUT24RC(3)025).

\bibliography{aaai25}

\end{document}